\documentclass{article}

\usepackage{arxiv}

\usepackage[normalem]{ulem}
\usepackage{graphicx}
\usepackage{amsmath}
\usepackage{makecell}
\usepackage{tikz}
\usepackage{multirow}
\usepackage{multicol}
\usepackage{tabu}

\title{Learning representations with end-to-end models for improved remaining useful life prognostic}

\author{
  Alaaeddine CHAOUB \\
  Université de Lorraine\\
    CNRS, LORIA, CRAN\\
  Nancy, France \\
  \texttt{alaaeddine.chaoub@loria.fr} \\
   \And
 Alexandre VOISIN \\
  Université de Lorraine\\
  CNRS, CRAN \\
  Nancy, France\\
  \texttt{alexandre.voisin@univ-lorraine.fr} \\
  \And
 Christophe CERISARA \\
  Université de Lorraine\\
  CNRS, INRIA, LORIA \\
  Nancy, France \\
  \texttt{christophe.cerisara@loria.fr} \\
  \And
 Benoît IUNG \\
  Université de Lorraine\\
  CNRS, CRAN\\
  Nancy, France\\
  \texttt{benoit.iung@univ-lorraine.fr} \\
  
}

\begin{document}
\maketitle
\begin{abstract}
\textcolor{black}{Remaining Useful Life (RUL) of equipment is defined as the duration between the current time and the time when it no longer performs its intended function. } An accurate and reliable prognostic of the remaining useful life provides decision-makers with valuable information to adopt an appropriate maintenance strategy to maximize equipment utilization and avoid costly breakdowns. In this work, we propose an end-to-end deep learning model based on multi-layer perceptron and long short-term memory layers (LSTM) to predict the RUL. After normalization of all data, inputs are fed directly to an MLP layers for feature learning, then to an LSTM layer to capture temporal dependencies, and finally to other MLP layers for RUL prognostic. The proposed architecture is tested on the NASA commercial modular aero-propulsion system simulation (C-MAPSS) dataset. Despite its simplicity with respect to other recently proposed models, the model developed outperforms them with a significant decrease in the competition score and in the root mean square error score between the predicted and the gold value of the RUL. In this paper, we will discuss how the proposed end-to-end model is able to achieve such good results and compare it to other deep learning and state-of-the-art methods.
\end{abstract}

\section{Introduction}
Maintenance is a crucial and costly activity: studies show that depending on the industry, between 15 and 70 percent of total production costs originate from maintenance activities \cite{krupitzer2020survey}. Nevertheless, maintenance has always been a main factor in a company's ability to be competitive in performance and deliver pricely and  high quality product. To that end, Prognostics and Health Management (PHM) is receiving a lot of attention in recent years thanks to its ability to drive maintenance in a more optimal way.
Prognostic or remaining useful life estimation, on top of diagnosis and fault detection, remains the \textcolor{black}{core} topic in PHM, as it provides PHM the ability to anticipate fault and provide relevant information on failure time to maintenance decision-makers.

\textcolor{black}{
Numerous prognostic algorithms for RUL estimation have been reported in the literature. The approaches for this challenging task can roughly be classified into two main approaches:
Model-based approaches \cite{cauchi2017model} \cite{yuan2013predictive}, and
Data driven approaches \cite{zhang2019data} .
The focus of this work is on RUL prognostic based on a data driven approach, more precisely deep learning models.}

\textcolor{black}{A classical way to tackle  prognostics, due to the variety of PHM problems and cases, relies on a first stage of feature selection or engineering before introducing them into machine learning (ML) models. One of the main feature of deep learning (DL) models is their ability to handle a large number of inputs and leverage complex correlation patterns among them. Surprisingly, when screening the literature, DL models used for prognostic are often preceded by such a feature selection stage. The originality of this work is to propose an end-to-end DL model that start with an MLP in the first layers for predicting the RUL using raw normalized input directly, an approach that reduce feature selection expenditure and deal with complex datasets with multiple operating conditions. We apply the model on the publicly available data sets C-MAPSS \cite{saxena2008damage} that describes the operational history of simulated aircraft turbofan engines. The result shows that, despite its simplicity, our model performs better than state-of-the-art approaches on this dataset.}

The rest of this paper is structured as follows. Section 2 introduces related works on RUL prognostic. Section 3 describes the proposed RUL prognostic architecture. Section 4 highlights the effectiveness of the proposed method by comparing the results with other popular methods. Finally, conclusions and discussion are provided in section 5.

\section{Related work}

RUL prognostic methods based on artificial intelligence approaches are attracting increasing attention, due to their ability to model highly nonlinear, complex and multidimensional systems. A number of deep learning (DL) techniques have been deployed in order to learn the mapping from monitored system data to their associated RUL. 
In this section, we focus on recent state-of-the-art work applying DL model to the C-MAPSS Dataset \cite{saxena2008damage}. For a general oveview, \cite{zhang2019review} review deep learning approaches applied to Prognostics and Health Management.

Recurrent neural networks are often used for problems involving time series data, because of their ability to process information over time.
\cite{zheng2017long} proposed a model based on multiple LSTM layers followed by a feed forward neural network that maps the input features to the predicted RUL, which is a standard deep learning architecture to deal with sequence data.
\cite{huang2019bidirectional} used a similar architecture but with Bidirectional LSTM cells in order to capture relevant information from both directions over time.

Convolutional neural networks are also often used when it comes to dealing with time series data thanks to their ability to model correlations in a temporal window around every time frame. \cite{li2020remaining} proposed an architecture based on multi-scale convolution kernels to capture information at different scales in the network, which helps to learn temporal features from different sequence sizes.

Trying to combine the advantages of different techniques,
a hybrid architecture was proposed in \cite{al2019hybrid}, integrating a deep LSTM and a deep CNN followed by a Multi-Layer Perceptron (MLP) to improve the prognostic performance.

Semi supervised learning was employed in \cite{hou2020remaining}, where deep convolutional generative adversarial networks are used.
The generator is an auto-encoder that tries to reconstruct the input signals, while the discriminator tries to distinguish the true data from the false ones. After this pre-training phase, the encoded features are used as input to an LSTM/MLP model for RUL prognostic.

These related works and other publications suggest that various deep learning architectures have been proposed and tested for RUL prognostic. Nevertheless, we argue that most of them may suffer from two potential issues. First, including a first stage of feature selection may harm the subsequent modeling process because it may genuinely discard relevant information and weak signals that may be hidden and overlooked by experts. Second, simple but well-designed neural networks often prove to match the performances of more complex deep learning architectures, whose hyper-parameters are more difficult to tune, eventually requiring many time-consuming and energy-hungry experiments, which presents a technical barrier for industrial applications.

\section{Proposed model architecture}
\textcolor{black}{To overcome the two aforementioned drawbacks, the proposed model has an MLP-LSTM-MLP architecture trained in an end-to-end manner for RUL prediction. This architecture has also been proposed in \cite{AN2020} and has shown promising results for diagnostic applications.}

Long Short Term Memory networks (LSTM) address the gradient vanishing problems in Vanilla recurrent networks by introducing new gates that allow for better control of gradient flow, and better preservation of long-term dependencies, which is needed in applications like RUL prognostic.
However, LSTM cells are designed to capture time dependencies but they do not have the capacity to handle complex feature processing, which has led other works in the literature to perform this task manually before the learning phase. Conversely, MLP are well fitted to perform such a task. We thus propose to feed all of the raw inputs into an MLP before the LSTM layers. The MLP will be in charge of processing the raw inputs and learning a good representation of each time frame, while the LSTM shall capture the dependencies through time of frame sequences. Then, a final regression head, composed of another MLP, predicts the RUL from these temporally smoothed representations.

\begin{figure*}[ht]
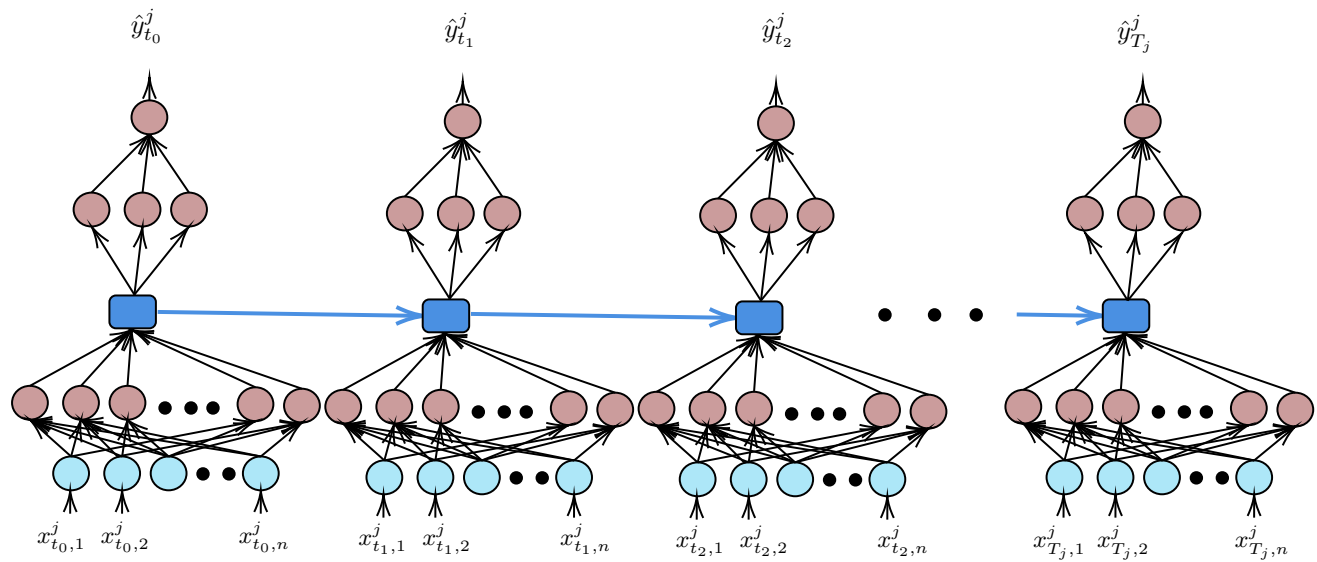

\centering

\tikzset{every picture/.style={line width=0.75pt}} 


\caption{Architecture of the proposed model: it takes as input a complete sequence $j$ of raw sensor values, encoded as a tensor $x^j$ composed of $T_{j}$ time frames with $n$-dimensional observations each. At training time, this sequence ranges from the first observed time frame $t_0$ to the last $T_j$ just before the Turbofan $j$ halts. At test time, a single forward pass is performed and the RUL $\hat y_{t}$ is predicted at every time step given the previous observations. To simplify the diagram, only one layer has been drawn for the MLPs.}
\label{model}
\end{figure*}

Figure \ref{model} shows the proposed architecture: each input vector \textcolor{black}{$x_t$} is processed by a first MLP with 3 layers, and the resulting sequence of feature vectors is processed by a single LSTM layer. The output of each LSTM cell is finally passed to another MLP with 3 layers that outputs a scalar \textcolor{black}{$y_t$} that represents the predicted RUL. The weights of the features-MLPs are shared across all time steps, which is convenient when working with variable length sequences.

\section{Experiments}
\subsection{The C-MAPSS Dataset overview}
The commercial modular aero-propulsion system simulation (C-MAPSS) is a turbofan engine simulation environment from NASA that provides access to health, control, and engine parameters through a graphical user interface (GUI). The C-MAPSS dataset \cite{saxena2008damage} is generated using the simulation program by monitoring the degradation of multiple Turbofan engines.

The data set is divided into four sub-data sets (from FD001 to FD004) with varying number of operating conditions and fault modes (see Table~(\ref{table:1})). Each sub-data set is further divided into training and test subsets. The training set is composed of input time series which are assumed to go on until failure. In the test set, time series are truncated arbitrarily and the objective is to estimate the number of remaining operational cycles before the system failure occurs.

The Turbofan Engine Degradation Simulation data sets (Table~(\ref{table:1})) are widely used by academics and scholars to test prognostic algorithms.
\begin{table}[ht]
\centering
\caption{The C-MAPSS Datasets}
{\renewcommand{\arraystretch}{2}
{\footnotesize
\begin{tabular}{| c | c | c | c | c | c|}
    \hline
    \multicolumn{2}{ |c| }{DATASET} & FD001 & FD002 & FD003 & FD004 \\ [1ex] 
    \hline\hline
    \multicolumn{2}{ |c| }{Nb Train trajectories} & 100 & 260 & 100 & 249  \\ 
    \hline
    \multicolumn{2}{ |c| }{Nb Test trajectories} & 100 & 259 & 100 & 249  \\
    \hline
    \multicolumn{2}{ |c| }{Nb Operating conditions} & 1 & 6 & 1 & 6  \\
    \hline
    \multicolumn{2}{ |c| }{Nb Fault modes} & 1 & 1 & 2 & 2  \\ 
    \hline
    \multirow{3}{*}{\shortstack{trajectories length\\ distribution}} & Max & 360 & 378 & 525 & 543
    \\
    \cline{2-6}
    & Mean & 206 & 206 & 247 & 245
    \\
    \cline{2-6}
    & Min & 128 & 128 & 145 & 128 
    \\
    \hline
    
\end{tabular}
}
}
\label{table:1}

\end{table}
\raggedbottom

The data contains multivariate time series, which correspond to 24 sensors measurements taken at each operating cycle of a particular simulated turbofan engine.

\subsection{Corpus preparation}

\begin{figure}[]
\centering
\includegraphics[scale=0.33]{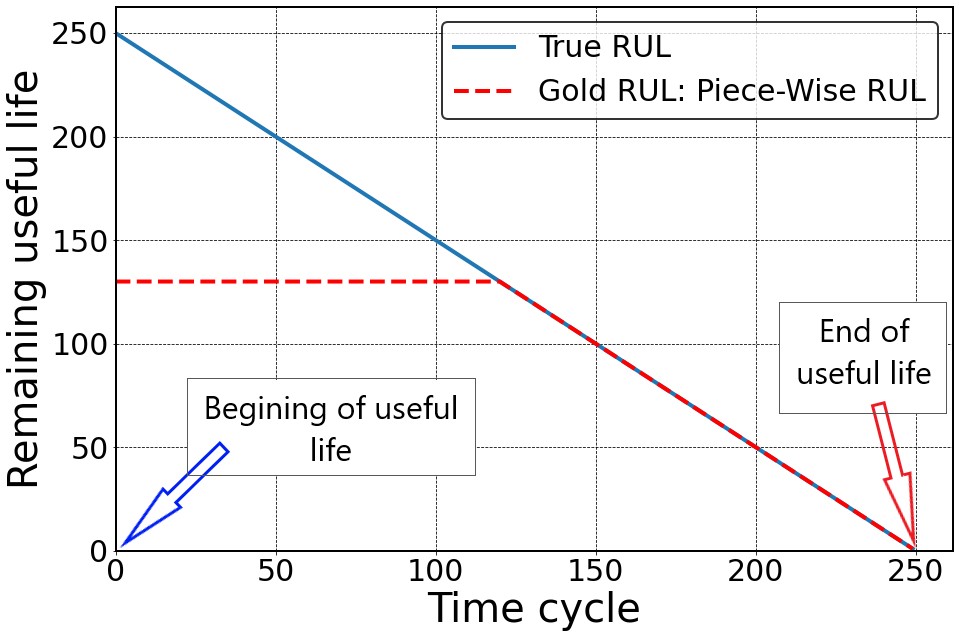}
\caption[]{True RUL vs Gold RUL of one trajectory with a length = 250 (Piece-wise maximum RUL is 130 time cycles).}
\label{piece-wise}
\end{figure}

\textcolor{black}{A gold RUL value for every cycle (or equivalently, time frame) is computed on the training set.
This is achieved by assuming that the RUL decreases linearly over time. \\
In practice, the degradation of a turbofan engine may be considered as negligible at the beginning of its use, and increases as the component approaches the end of its life.
To better model the changes in the remaining useful life as function of time with respect to the non linearity of the degradation, we adopt a strategy that is often used in related works, and model the RUL with a piece-wise linear function shown in Figure~(\ref{piece-wise}), which limits the maximum RUL to a constant value and then begins linear degradation after a certain time \cite{RUL2008}. The most common maximal RUL values used among the works in the literature are 125 and 130 \cite{al2019hybrid} \cite{zheng2017long}. In the following, we choose a maximum RUL of 130.\\
The gold RUL is defined in Eq. \ref{gold rul} as:
\begin{equation}
\begin{array}{l}
Gold~RUL = 
\begin{cases} 
130 & \mathrm {if}~True~RUL \geq 130 \\ 
True~RUL & \mathrm {if}~True~RUL < 130
\end{cases} 
\label{gold rul}
\end{array}
\end{equation}
}

\textcolor{black}{This maximum value to predict shall facilitate training the model, because it reduces the range of useful values to predict. However, it may also prevent the model from being able to predict very long RUL values, but we expect this case to rarely occur in the test set.}

Another constraint on data comes from the choice of a deep learning model, which trains better when all inputs and outputs are normalized.
Hence, we normalize both inputs and outputs in the range $[0,1]$ using Equation~\ref{eq1} for the inputs, and similarly for the RUL values:
\begin{equation}
\label{eq1}
x^{j}_{t,i} =\frac{v^{j}_{t,i}-\min_{t,j}(v^{j}_{t,i})}{\max_{t,j}(v^{j}_{t,i})-\min_{t,j}(v^{j}_{t,i})}
\end{equation}

Where $v^{j}_{t,i}$ is the value of the $i_{th}$ sensor at time $t$ from engine $j$, and $x^{j}_{t,i}$ is the corresponding normalized value. In order to compute the min and max values, $t$ and $j$ range across the entire data set, i.e. all trajectories of all turbofans are used.
Normalization helps to stabilize training of the network parameters, speeds up convergence of gradient descent and reduces the risk of getting stuck in local optima.

The normalized data is directly fed to the network, without any feature engineering or selection. Therefore, no prior expertise on Turbofans or signal processing is required for the proposed method.

\subsection{Performance metrics}
In the PHM context, it is generally desirable to predict failures as early as possible. Therefore, \textcolor{black}{the scoring function that will be used to evaluate the performances of the models penalizes more the errors that predict a RUL too late than too early,and it is given by Eq.~\ref{score} as proposed in the original C-MAPSS evaluation campaign~\cite{saxena2008damage}:}

\begin{equation}
\begin{array}{l}

S =\sum^{N}_{j=1}\sum^{T_{j}}_{t=1} s^{j}_{t} \text{ , Where: } \\ \\
s^{j}_{t}= 
\begin{cases} 
{e}^{-\frac {d^{j}_{t}}{13}}-1 & \mathrm {for}~d^{j}_{t}<0 \\ 
{e}^{\frac {d^{j}_{t}}{10}}-1 & \mathrm {for}~d^{j}_{t}\geq 0
\end{cases} 
\label{score}
\end{array}
\end{equation}

Where  $N$,$T_{j}$ are respectively the number, length of the trajectories, and $d^{j}_{t} = \hat y^{j}_{t}-y^{j}_{t}$ (Predicted RUL - Gold RUL)

For the sake of comparability with other literature results, we will use the scoring function of the challenge plus the Root Mean Square Error (RMSE) (Eq. (\ref{rmse})):
\begin{equation}
RMSE = \sqrt{\frac{1}{N} \sum^{N}_{j=1}\frac 1 {T_j} \sum^{T_{j}}_{t=1}{d^{j}_{t}}^2}
\label{rmse}
\end{equation}

\begin{figure}[h]
\centering
\includegraphics[scale=0.9]{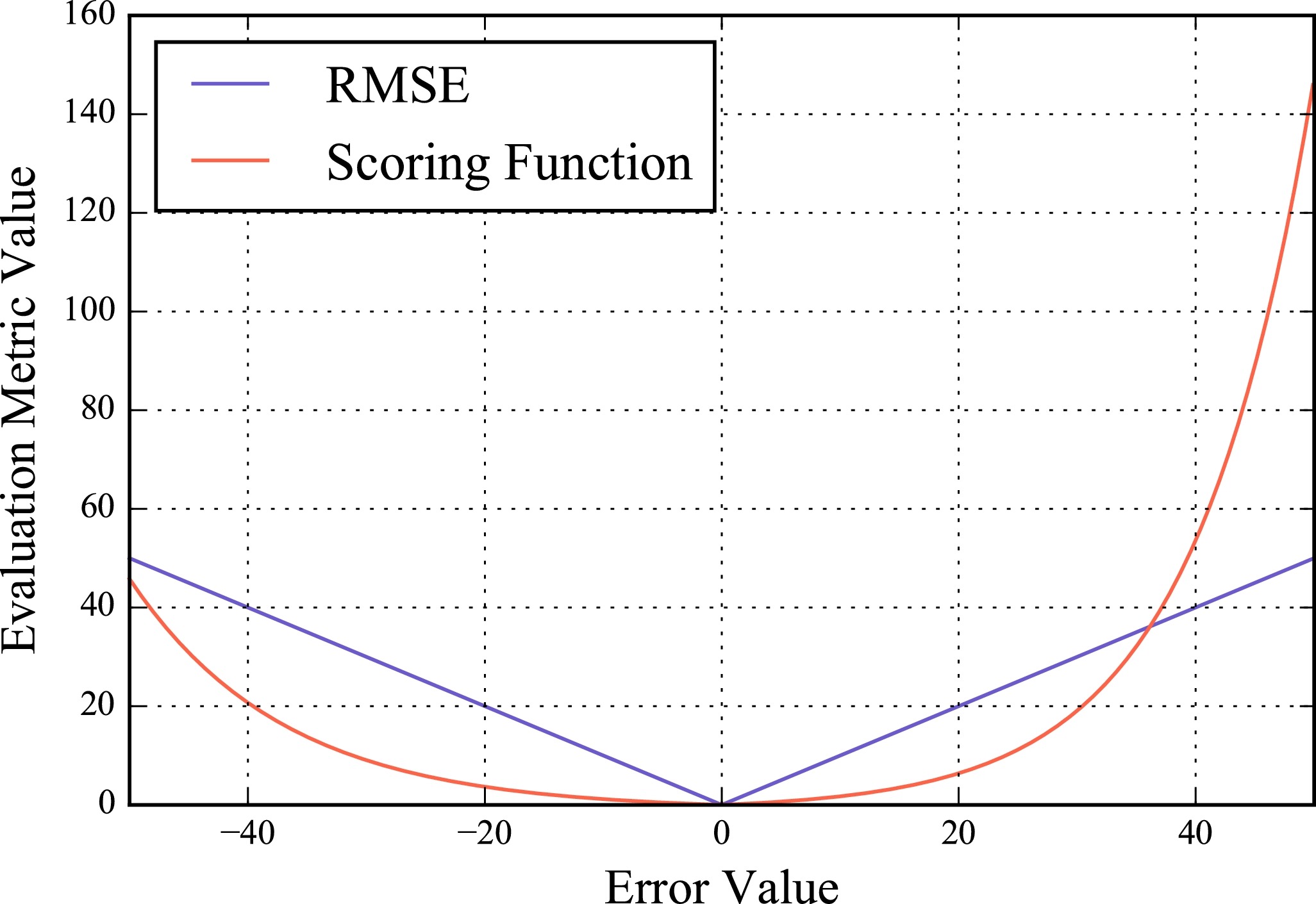}
\caption[]{Comparison between the scoring function and RMSE with respect to different error values.}
\label{losses}
\end{figure}

Figure (\ref{losses}) shows how both metrics penalize errors in detail. However, the main objective is to achieve the smallest value possible for both.

\begin{figure*}
    \centering

   \includegraphics[scale= 0.2]{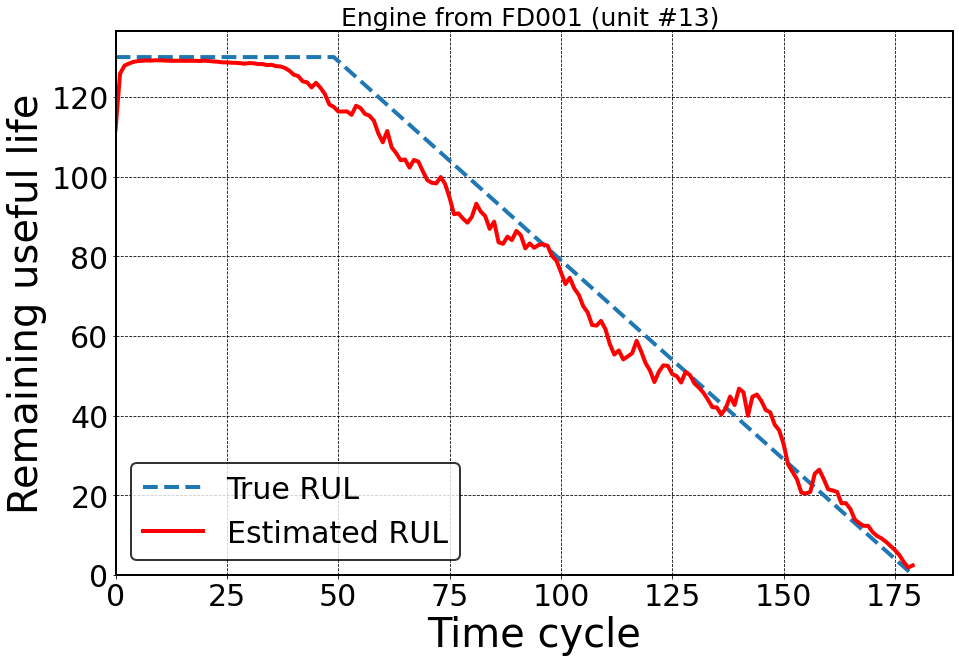}
    \hspace*{0pt}
    \includegraphics[scale= 0.2]{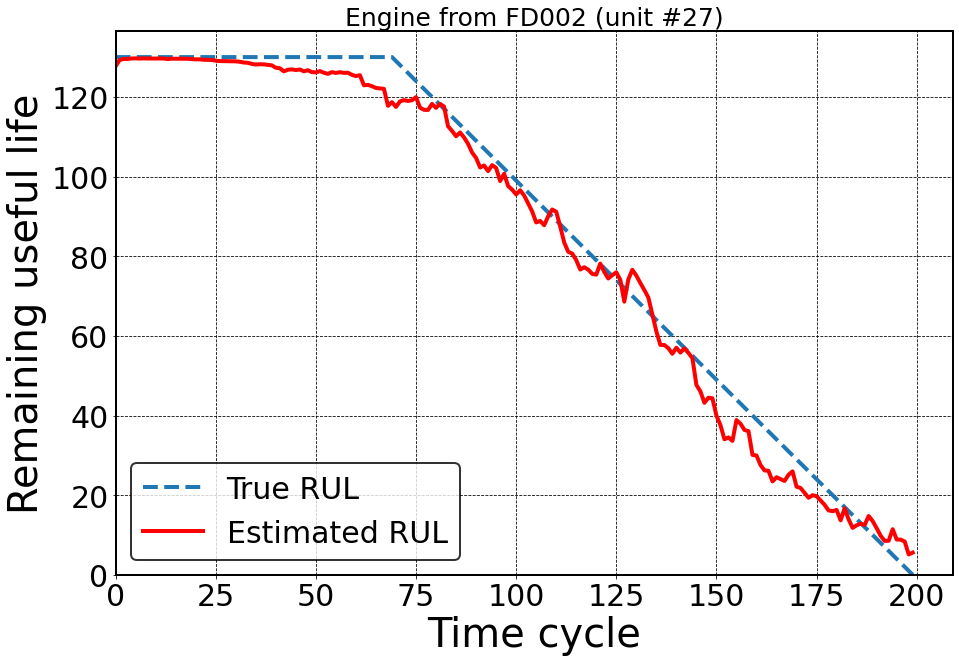}
    \hspace*{0pt}
    \includegraphics[scale= 0.2]{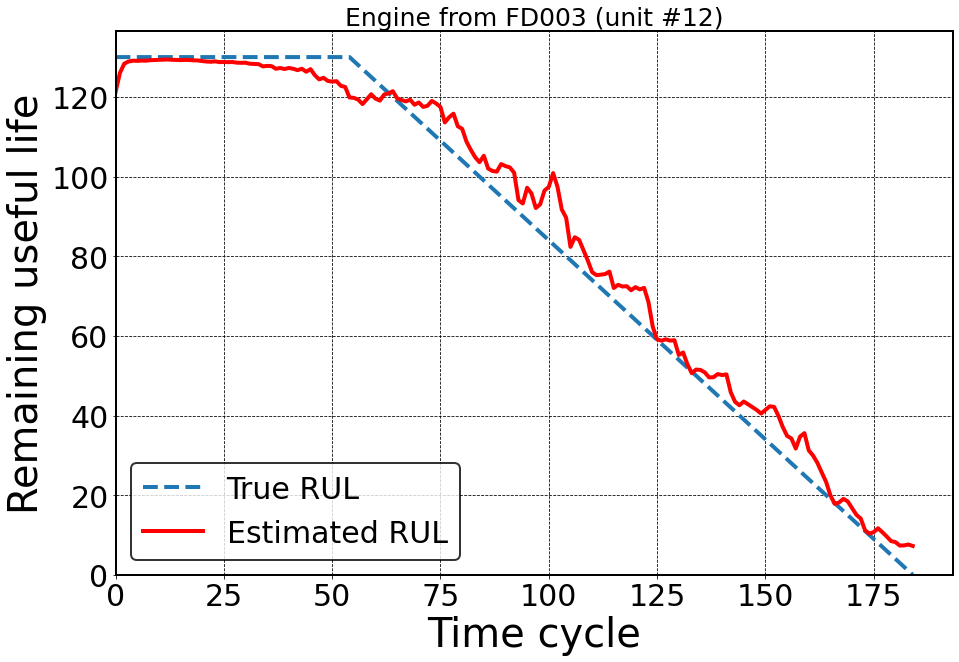}
    \hspace*{0pt}
    \includegraphics[scale= 0.2]{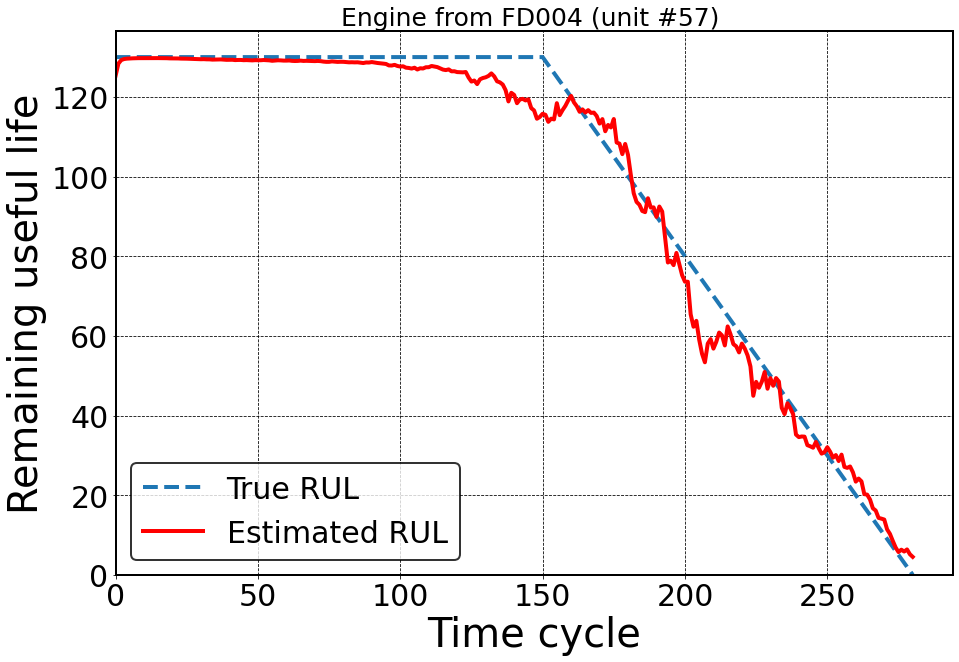}
    \hspace*{0pt}
    \caption{Different examples of RUL prognostic for engine units from the validation samples of each data set. }
    \label{pred_4}
\end{figure*}

\subsection{Model Training and evaluation}
In order to choose the hyper-parameters of the model, we split the training set into a training subset and a validation subset, based on the ID of the engines. The original test set is reserved for final evaluation.

Our model does not use neither fixed length sequences nor truncation nor padding, and each training sample is a full time series of one turbofan engine from its first cycle until failure. Henceforth, different samples have variable sequence length.
We used 75\% of the turbofans run to failure trajectories as training subset, and 25\% as validation subset.

Both metrics may be used as loss functions for training. Preliminary experiments show that both the score and the RMSE give similar results. So we decided to work with the RMSE because the training process is faster.

Hyper-parameters have been tuned manually with a few trials and errors on the validation set. The hyper-parameters to be optimized are the learning rate, the number of layers in the input and output MLPs, the number of LSTM layers, the number of neurons/cells in each layer, the activation functions, the dropout percentages and the optimizer. The best hyper-parameters found for the proposed model are listed in Table~(\ref{table:2}).

\begin{table}[ht]
\begin{center}
\caption{Hyper-parameters of the proposed model}
{\footnotesize
\begin{tabular}{ c  c   }
    \hline
    Hyper-parameter & Value \\
    \hline \\
    Learning Rate  & 0.0001   \\[1ex] 

    Number of MLP layers before LSTM  & 3   \\[1ex] 
    
    Number of neurons in MLP layers & 100/50/50   \\[1ex] 
    
    Number of LSTM layers & 1  \\[1ex] 
    
    Number of LSTM cells  & 60  \\ [1ex]
    
    Number of MLP layers after LSTM  & 3   \\[1ex] 
    
    Number of neurons in MLP layers & 60/30/1   \\[1ex] 
    
    Activation function for MLP layers & Tanh()   \\[1ex]
    
    Batch size & 5   \\[1ex]
    
    Dropout percentage & 0\%   \\[1ex]
    \hline
\end{tabular}
}
\label{table:2}
\end{center}
\end{table}

\begin{figure*}[ht]
\centering

    \includegraphics[scale = 0.3]{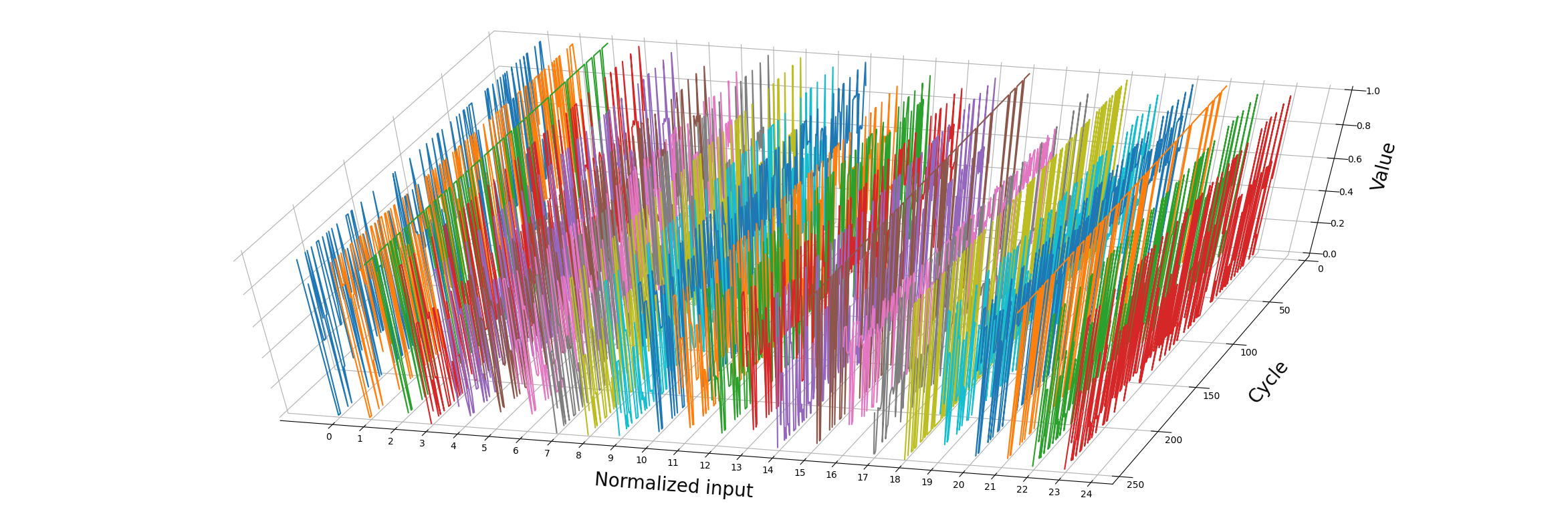}
    \caption{
    Normalized input signals that are fed directly into the model; $n=24$ sensor measurements of the turbofan unit \#13 from the beginning of its life until its failure; this engine data was taken from the $4^{th}$ data set (FD004) that contains 6 operating conditions and 2 fault modes; we clearly see that these normalized signals do not directly provide visible and interpretable clues for RUL estimation.}
    \label{FD004_input}

   \includegraphics[scale = 0.3]{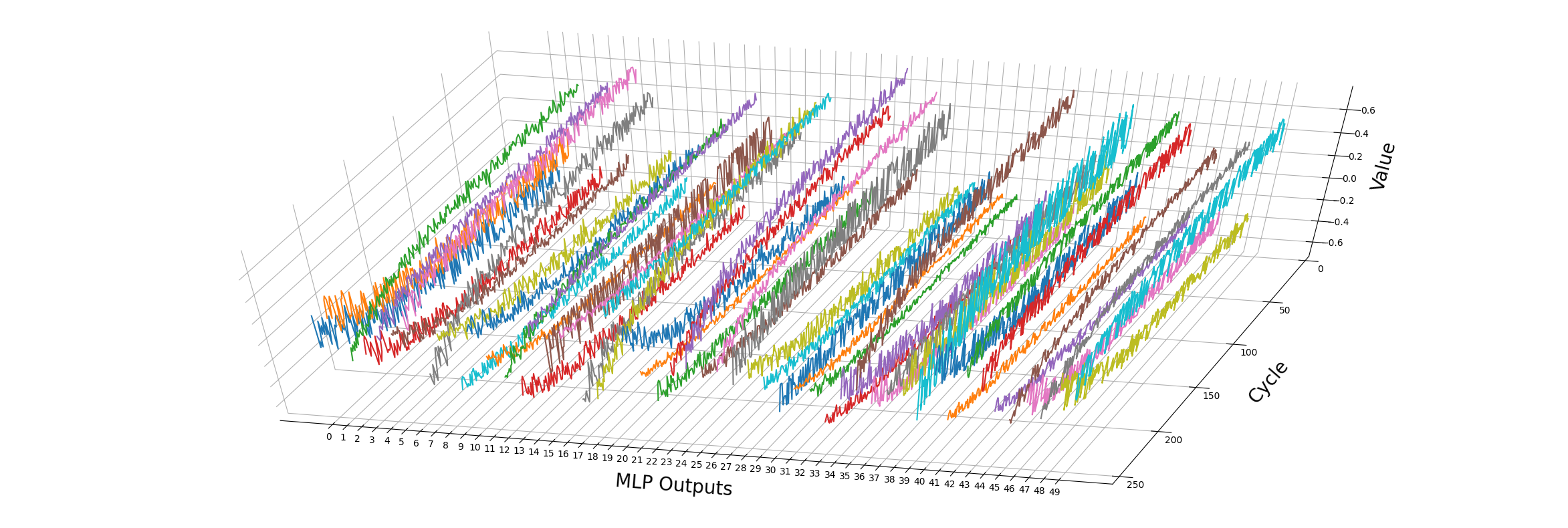}
    \caption{This plot presents the 50 features learned by the MLP for unit \#13; we can observe trending degradation representations that have been learned from the normalized input signals. Since the first MLP is not time dependent, the learned features exhibit a relatively large variance across time cycles.}
    \label{FD004_MLP}

    \includegraphics[scale = 0.3]{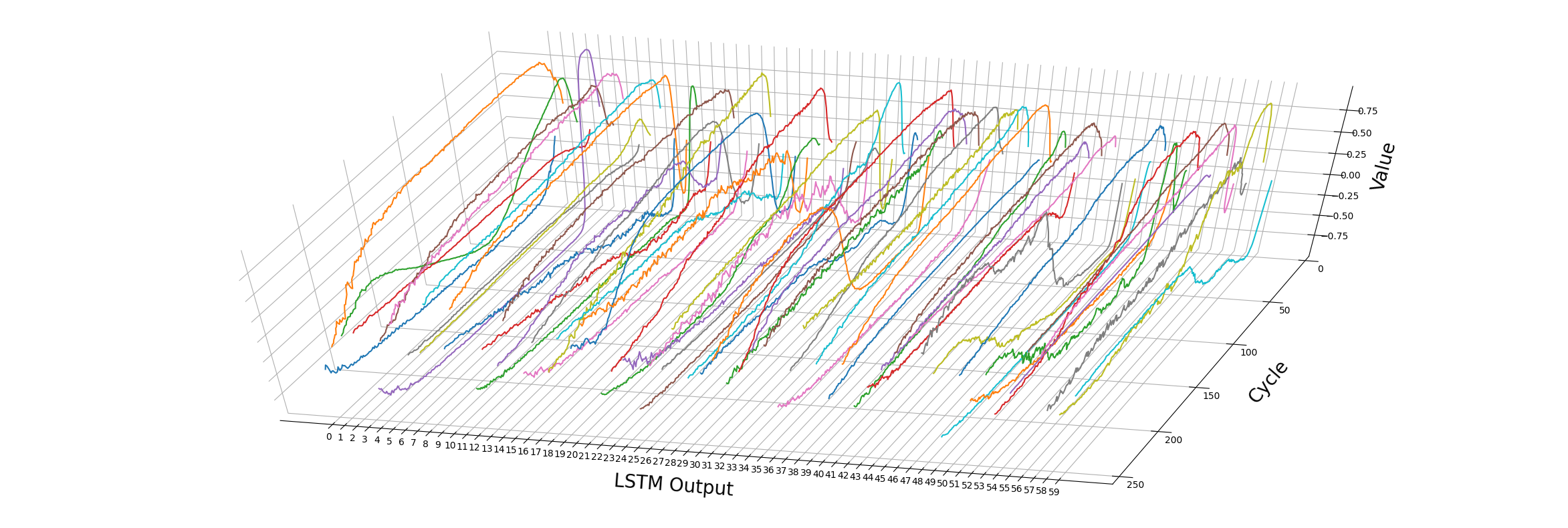}
    \caption{The outputs of the LSTM for unit \#13 present much smother signals, due to the LSTM's ability to leverage recurrent connection from prior time steps.}
    \label{FD004_lstm}

\end{figure*}

\subsection{Results and Discussion}
Because of random initialization, the optimized model parameters values may vary across different training runs. We thus evaluate the model's performances across 10 runs, and the mean values and standard deviations are given in Table~(\ref{table:3}) for the four data sets.

\begin{table}[ht!]
\centering
\caption{prognostic performance of the proposed model.}
{\renewcommand{\arraystretch}{2.5}
{\footnotesize
\begin{tabular}{| c | c | c | c | c|}
    \hline
    
    DATASET & FD001 & FD002 & FD003 & FD004 \\ [1ex] 
    \hline\hline
    \makecell{RMSE}
    & \makecell{13.26 \\ $\pm$ 0.57} 
    & \makecell{12.49 \\ $\pm$ 0.28}
    & \makecell{13.11 \\ $\pm$ 1.28}
    & \makecell{13.97 \\ $\pm$ 0.48}  \\ 
    \hline \makecell{SCORE}
    & \makecell{284.88 \\ $\pm$ 42.32 }
    & \makecell{571.4 \\ $\pm$ 37.45 }
    & \makecell{352.39 \\ $\pm$ 179.96 }
    & \makecell{1252.32 \\ $\pm$ 104.97 }\\

    \hline
\end{tabular}
}
}
\label{table:3}
\end{table}

Figure~(\ref{pred_4}) shows the predicted RUL vs the gold value of the RUL for four trajectories from the validation subset. We see that the proposed model can follow degradation patterns even in complex data sets as FD002 and FD004 with 6 operating conditions. Thanks to our end to end learning approach, the MLP that precedes the LSTM automatically learns a representation of the input data that is relevant to the task of RUL prediction. Figure~(\ref{FD004_input}) shows the normalized raw input signals of unit \#13 from the FD004 data set, where no clear trend can be seen because of the high variance in the data, which is partly due to the operating conditions that vary from cycle to cycle.
Figure~(\ref{FD004_MLP}) shows the output signals of the first MLP, where noticeable degradation pattern have been learned from the normalized inputs and can be observed.
Feeding this learned representation to the rest of the model is more efficient than handcrafting features that require expertise and time.
This first representation learning stage is particularly useful when dealing with complex data sets where no clear trend is seen, and also when inputs have a large number of dimensions.

After this first MLP, the role of the LSTM layer is to capture temporal patterns and dependencies in the time series. Figure~(\ref{FD004_lstm}) shows the signal at the
output of the LSTM. We can see that this part of the model minimizes the variance of the learned features across time cycles giving a smoother signals that can be used by the final MLP for RUL estimation.

\subsection{Comparison with related works}
We evaluate in Table~(\ref{comparison}) our proposed model by comparing its performances with the most recent methods published in the literature that give the best results on the C-MAPSS data set to the best of our knowledge.

\begin{table}[ht!]
\centering

\caption[]{Performance comparison of related methods with our proposed model on the C-MAPSS benchmark.}
{\renewcommand{\arraystretch}{2.5}
\scalebox{0.78}{
\begin{tabu}{|l| c | c | c | c | c | c | c | c || c | c |}

    \hline
    
   \multirow{3}{*}{Models} & \multicolumn{8}{ c|| }{DATA SETS}&
   \multicolumn{2}{ c| }{}\\
    \cline{2-9}
    
     & \multicolumn{2}{ c| }{FD001} & \multicolumn{2}{ c| }{FD002} & \multicolumn{2}{ c| }{FD003} & \multicolumn{2}{ c|| }{FD004}& 
     \multicolumn{2}{ c| }{{\bf Average}}\\
    \cline{2-11}
    
    & \makecell{RMSE} 
    & \makecell{Score}
    & \makecell{RMSE}
    & \makecell{Score}
    & \makecell{RMSE}
    & \makecell{Score}
    & \makecell{RMSE}
    & \makecell{Score}
    & \makecell{RMSE}
    & \makecell{Score}\\
    \hline
    \hline
    
    DA-CNN \cite{song2020distributed}
    & \makecell{11.78}
    & \makecell{229.48}
    & \makecell{16.95}
    & \makecell{1842.38}
    & \makecell{11.56}
    & \makecell{257.11}
    & \makecell{18.23}
    & \makecell{2317.32}
    & \makecell{14.63 }
    & \makecell{1161.57 }\\
    \hline
    
    DCGAN \cite{hou2020remaining}
    & \makecell{\textbf{10.71}}
    & \makecell{\textbf{174}}
    & \makecell{19.49}
    & \makecell{2982}
    & \makecell{\textbf{11.48}}
    & \makecell{273}
    & \makecell{19.71}
    & \makecell{3874}
     & \makecell{15.34 }
     & \makecell{1825.75 }\\

    \hline
    
    MS-DCNN \cite{li2020remaining}
    & \makecell{11.44}
    & \makecell{196.22}
    & \makecell{19.35}
    & \makecell{3747}
    & \makecell{11.67}
    & \makecell{\textbf{241.89}}
    & \makecell{22.22}
    & \makecell{4844}
     & \makecell{16.17 }
     & \makecell{2257.27 }\\
    
    \hline
    
    HDNN \cite{al2019hybrid}
    & 13.017
    & 245
    & 15.24
    & \makecell{1282.42}
    & \makecell{12.22}
    & \makecell{287.72}
    & \makecell{18.15}
    & \makecell{1527.42}
    & \makecell{14.65 }
    & \makecell{835.64 }\\
    
    \hline 
    \rowfont{\color{black}}
    LSTM \cite{Pasa2019}
    & \makecell{16.5}
    & \makecell{444}
    & \makecell{18.1}
    & \makecell{942}
    & \makecell{15.9}
    & \makecell{718}
    & \makecell{17.2}
    & \makecell{1487}
    & \makecell{16.92}
    & \makecell{897.75}\\
    
    \hline
    \rowfont{\color{black}}
    \textbf{Proposed LSTM without the first MLP}
    
    & \makecell{14.31}
    & \makecell{337.86}
    & \makecell{17.44}
    & \makecell{1716.11}
    & \makecell{15.53}
    & \makecell{1356.36}
    & \makecell{18.86}
    & \makecell{2111.05}
    & \makecell{16.53}
    & \makecell{1380.34}\\
    
    \hline
    
    \textbf{Proposed LSTM with the first MLP}
    & \makecell{13.26 }
    & \makecell{284.88 }
    & \makecell{\textbf{12.49} }
    & \makecell{\textbf{571.4} }
    & \makecell{13.11 }
    & \makecell{352.39 }
    & \makecell{\textbf{13.97} }
    & \makecell{\textbf{1252.32} }
     & \makecell{\textbf{13.20} }
     & \makecell{\textbf{615.24} }\\
    \hline
\end{tabu}
}
}
\label{comparison}

\end{table}

\textcolor{black}{Although the previously published models are performing well on the first and third data sets (FD001, FD003), with only one operating condition, they perform poorly on the other subsets that have up to 6 operating conditions, except for the approaches proposed in \cite{al2019hybrid} and in \cite{Pasa2019}.\\
The proposed end-to-end architecture outperforms all other models in complex data sets (FD002 and FD004) as well as on the global results averaged over all datasets.
It improves by more than 18\% for the RMSE and 39\% for the Score on FD002,
and 18\% for the RMSE and 15 \% for the Score on the FD004 data set,
as compared to literature results.\\
We explain these good results as a consequence of adding a first representation learning MLP before the LSTM and training both of them in an end-to-end manner.
Indeed, we can see from the last two rows of Table~(\ref{comparison}) that the results improved significantly after adding the first MLP to the architecture.\\
The outputs of the first MLP shows that this part is removing a large part of the variability of the sensor signals that is due to varying operating conditions (Figure~(\ref{FD004_MLP})). This greatly facilitates the work of the LSTM that can focus on temporal smoothing, and then of the final MLP, which role is to achieve prediction. This idea of facilitating the work of the LSTM can also be achieved by feature engineering as proposed in \cite{Pasa2019}, where they normalized input according to the operating conditions, the results are relatively good, but this approach can not be performed when some or all of the operating conditions are not known, unlike the proposed approach in this work.\\
The clear decomposition in our model of these three roles is the key to the increased robustness to variable input signals and better final performances.
This can be also observed in Table~(\ref{comparison}), where all competing models suffer from a large variability of their performance between the FD001 and FD003 subsets on the one hand, and the FD002 and FD004 subsets on
the other hand, while a significantly lower difference in the results between the 4 subsets can be observed with our proposed model.}

\section{Conclusion and future work}

In this paper, we presented an end-to-end deep learning approach for RUL estimation from multivariate time-series signals. 
The proposed method has been tested on the public C-MAPSS data set where the goal is to predict the RUL of commercial aero-engine units. Comparisons with several state-of-the-art approaches have been conducted. \textcolor{black}{The results show that our proposed neural architecture gives the best scores when compared to other approaches applied to the same data sets, }especially on complex ones with different operating conditions. Furthermore, it exhibits a more consistent behaviour across the four datasets.

\textcolor{black}{In future work, we plan to explore the performances of the proposed model on other more realistic data sets with different input size, where we may not be able to use the same approach as used in this paper, such as feeding the entire run to failure trajectories to the model due to their long length.}

\section*{Funding}
This work is part of the project AI-PROFICIENT which has received funding from the European Union’s Horizon 2020 research and innovation program under grant agreement No 957391. The Grid5000 computing resources have been used to partly train and evaluate the proposed models.

\bibliographystyle{unsrt}  

\bibliography{references.bib}

\begin{thebibliography}{10}

\bibitem{krupitzer2020survey}
Christian Krupitzer, Tim Wagenhals, Marwin Züfle, Veronika Lesch, Dominik
  Schäfer, Amin Mozaffarin, Janick Edinger, Christian Becker, and Samuel
  Kounev.
\newblock A survey on predictive maintenance for industry 4.0, 2020.

\bibitem{cauchi2017model}
Nathalie Cauchi, Karel Macek, and Alessandro Abate.
\newblock Model-based predictive maintenance in building automation systems
  with user discomfort.
\newblock {\em Energy}, 138:306--315, 2017.

\bibitem{yuan2013predictive}
Yong Yuan, Xiaomo Jiang, and Xian Liu.
\newblock Predictive maintenance of shield tunnels.
\newblock {\em Tunnelling and Underground Space Technology}, 38:69--86, 2013.

\bibitem{zhang2019data}
Weiting Zhang, Dong Yang, and Hongchao Wang.
\newblock Data-driven methods for predictive maintenance of industrial
  equipment: a survey.
\newblock {\em IEEE Systems Journal}, 13(3):2213--2227, 2019.

\bibitem{saxena2008damage}
Abhinav Saxena, Kai Goebel, Don Simon, and Neil Eklund.
\newblock Damage propagation modeling for aircraft engine run-to-failure
  simulation.
\newblock In {\em 2008 international conference on prognostics and health
  management}, pages 1--9. IEEE, 2008.

\bibitem{zhang2019review}
Liangwei Zhang, Jing Lin, Bin Liu, Zhicong Zhang, Xiaohui Yan, and Muheng Wei.
\newblock A review on deep learning applications in prognostics and health
  management.
\newblock {\em IEEE Access}, 7:162415--162438, 2019.

\bibitem{zheng2017long}
Shuai Zheng, Kosta Ristovski, Ahmed Farahat, and Chetan Gupta.
\newblock Long short-term memory network for remaining useful life estimation.
\newblock In {\em 2017 IEEE international conference on prognostics and health
  management (ICPHM)}, pages 88--95. IEEE, 2017.

\bibitem{huang2019bidirectional}
Cheng-Geng Huang, Hong-Zhong Huang, and Yan-Feng Li.
\newblock A bidirectional lstm prognostics method under multiple operational
  conditions.
\newblock {\em IEEE Transactions on Industrial Electronics}, 66(11):8792--8802,
  2019.

\bibitem{li2020remaining}
Han Li, Wei Zhao, Yuxi Zhang, and Enrico Zio.
\newblock Remaining useful life prediction using multi-scale deep convolutional
  neural network.
\newblock {\em Applied Soft Computing}, 89:106113, 2020.

\bibitem{al2019hybrid}
Ali Al-Dulaimi, Soheil Zabihi, Amir Asif, and Arash Mohammadi.
\newblock Hybrid deep neural network model for remaining useful life
  estimation.
\newblock In {\em ICASSP 2019-2019 IEEE International Conference on Acoustics
  Speech and Signal Processing (ICASSP)}, pages 3872--3876. IEEE, 2019.

\bibitem{hou2020remaining}
Guisheng Hou, Shuo Xu, Nan Zhou, Lei Yang, and Quanhao Fu.
\newblock Remaining useful life estimation using deep convolutional generative
  adversarial networks based on an autoencoder scheme.
\newblock {\em Computational Intelligence and Neuroscience}, 2020, 2020.

\bibitem{AN2020}
Zenghui An, Shunming Li, Jinrui Wang, and Xingxing Jiang.
\newblock A novel bearing intelligent fault diagnosis framework under
  time-varying working conditions using recurrent neural network.
\newblock {\em ISA Transactions}, 100:155--170, 2020.

\bibitem{RUL2008}
Felix~O. Heimes.
\newblock Recurrent neural networks for remaining useful life estimation.
\newblock In {\em 2008 International Conference on Prognostics and Health
  Management}, pages 1--6, 2008.

\bibitem{song2020distributed}
Yan Song, Shengyao Gao, Yibin Li, Lei Jia, Qiqiang Li, and Fuzhen Pang.
\newblock Distributed attention-based temporal convolutional network for
  remaining useful life prediction.
\newblock {\em IEEE Internet of Things Journal}, 2020.

\bibitem{Pasa2019}
Gabriel~Duarte Pasa, I.~Medeiros, and T.~Yoneyama.
\newblock Operating condition-invariant neural network-based prognostics
  methods applied on turbofan aircraft engines.
\newblock 2019.

\end{thebibliography}

\end{document}